\documentclass{article}

\PassOptionsToPackage{numbers, compress}{natbib}

\usepackage[preprint]{neurips_2025}

\usepackage[utf8]{inputenc} % allow utf-8 input
\usepackage[T1]{fontenc}    % use 8-bit T1 fonts
\usepackage{hyperref}       % hyperlinks
\usepackage{url}            % simple URL typesetting
\usepackage{booktabs}       % professional-quality tables
\usepackage{amsfonts}       % blackboard math symbols
\usepackage{nicefrac}       % compact symbols for 1/2, etc.
\usepackage{microtype}      % microtypography
\usepackage{xcolor}         % colors

\usepackage{epsfig}
\usepackage{graphicx}
\usepackage{amsmath}
\usepackage{amssymb}
\usepackage{enumitem}
\usepackage{multirow}
\usepackage{rotating}
\usepackage{booktabs}

\usepackage{float}
\usepackage{caption}
\usepackage{subcaption}
\usepackage{threeparttable}

\usepackage{pifont}% http://ctan.org/pkg/pifont

\usepackage{hyperref}
\usepackage{listings}

\title{Rooms from Motion: Un-posed Indoor 3D Object Detection as Localization and Mapping}

\author{%
  Justin Lazarow \quad Kai Kang \quad Afshin Dehghan\\
  Apple\\
  \texttt{\{jlazarow, kai\_kang, adehghan\}@apple.com} \\
}

\begin{document}

\maketitle

\begin{center}
    \centering
    \captionsetup{type=figure, labelformat=empty}
    \vspace{-2.em}
    \includegraphics[width=0.98\textwidth]{fig/small/new-teaser-small.pdf}
    \captionof{figure}{\footnotesize
    \textbf{Rooms from Motion} realizes an object-centric framework for metric localization and semantic 3D object-level mapping from \textit{un-posed} RGB images without the need for explicit 2D keypoints or point clouds. Given an unordered collection of images, Rooms from Motion detects every object as a metric 3D box within each image, uses a learned object matcher to associate objects across frames, estimates relative poses using the 3D boxes of matched objects, and finally estimates absolute camera poses and forms global, semantic 3D object tracks (akin to 3D object detection). \textbf{Above:} We visualize the semantics-aware map and camera localization of Rooms from Motion from un-posed RGB images on two challenging ScanNet++ scenes: a large laboratory and a few rooms within a residential space. \textbf{Below:} We show class-agnostic results on an open space from the CA-1M dataset.
    }
    \label{fig:teaser}
\end{center}

\begin{abstract}
    \vspace{-.5em}
    We revisit scene-level 3D object detection as the output of an object-centric framework capable of both localization and mapping using 3D oriented boxes as the underlying geometric primitive. While existing 3D object detection approaches operate globally and implicitly rely on the \textit{a priori} existence of metric camera poses, our method, \textbf{Rooms from Motion} (RfM) operates on a collection of un-posed images. By replacing the standard 2D keypoint-based matcher of structure-from-motion with an object-centric matcher based on image-derived 3D boxes, we estimate metric camera poses, object tracks, and finally produce a global, semantic 3D object map. When \textit{a priori} pose is available, we can significantly improve map quality through optimization of global 3D boxes against individual observations. RfM shows strong localization performance and subsequently produces maps of higher quality than leading point-based and multi-view 3D object detection methods on CA-1M and ScanNet++, despite these global methods relying on overparameterization through point clouds or dense volumes. Rooms from Motion achieves a general, object-centric representation which not only extends the work of Cubify Anything to full scenes but also allows for inherently sparse localization and parametric mapping proportional to the \textit{number of objects} in a scene.
\end{abstract}

\section{Introduction}

\textit{Can objects serve as the basis for localization and mapping?} Currently, \textit{points} are the fundamental primitive for localization and mapping techniques. For localization, this usually means matching keypoints between images. For mapping, this means producing sparse point-based representations. Whether these representations are subsequently densified, meshed, or voxelized, they lack explicit semantic and object-level understanding of the mapped spaces. Rather, it has been the goal of additional machinery (i.e., entire areas of research) to subsequently extract meaningful semantics and objects from the estimated point representations. However, if localization and an object-level representation of the world is desired, we set out to answer whether an object-level representation itself is sufficient for mapping and localization algorithms.

\textit{Can we effectively extract global 3D object-level structure without explicit point clouds or voxel grids?} Methods for global \cite{rukhovich2022fcaf3d, kolodiazhnyi2024unidet3d} or multi-view scene-level 3D object detection \cite{tu2023imgeonet, rukhovich2022imvoxelnet} require building an \textit{explicit} global 3D representation e.g., a point cloud or voxel grid. Not only is this an \textit{overparameterization} of what we want to extract (3D boxes), but it requires metric camera poses \textit{a priori}. However, having access to camera poses implies that a great deal of the 3D world has already been solved for. Therefore, we seek an alternative to merely creating more accurate 3D point-based representations from which objects may be extracted and instead seek to make the \textit{object} the central representation of 3D.

Towards answering these questions, we present a framework, \textbf{Rooms from Motion}, which allows for localization \textit{and} mapping within indoor spaces (i.e., rooms) using 3D objects as the fundamental primitive without the need for point-based representations, \textit{a priori} poses, or any metric quantities. We do this by constructing object-centric components analogous to existing point-based structure-from-motion principles. Given an unordered collection of RGB images, Rooms from Motion:

\begin{enumerate}
    \item Independently detects every object within each image in a metrically accurate way in the form of a oriented 3D bounding box alongside an embedding vector \cite{lazarow2024cubify}.
    \item  Given two frames, each with a set of detected 3D objects, a learned matching network determines object-level correspondences between the frames. Metric relative poses are derived using the set of matched 3D boxes.
    \item From the set of relative poses, traditional averaging components are used to form a global set of poses \cite{pan2024global, pan2024gravity}.
    \item 3D object tracks are established using the set of all matched 3D object pairs and ultimately produce a global 3D object map.
    \item Finally, global 3D objects can be refined using optimization to overcome inherent partial observations of objects.
\end{enumerate}

Rooms from Motion achieves this using both an internal and output representation defined by oriented 3D boxes. \textit{No larger representation is ever used}. Therefore, unlike volume or point-based methods which must explicitly build a world-space 3D representation, Rooms from Motion naturally scales to large scenes with a representation that scales proportional to the number of objects in a scene versus the actual geometry of the scene. Since the number of objects in a scene is dramatically smaller than any spatial quantity (e.g., voxels, distance fields, radiance fields) while still providing a continuous parametrization of each object, Rooms from Motion provides a compact, yet rich representation for downstream tasks.

\section{Related Work}

\noindent\textbf{Object-centric Localization and Mapping} Using objects as a basis for localization and mapping has appeared throughout the literature. SLAM++ \cite{salas2013slam++} considered the concept of ``object-oriented SLAM'' which operates on RGB-D measurements and localizes while mapping a set of pre-defined objects characterized by scanned meshes. Other representations like quadrics \cite{nicholson2018quadricslam, wu2023object, li2021odam} are often used to optimize multi-view consistency across 2D object detections. Rooms from Motion, however, is a generalist framework capable of mapping any object directly at the 3D box level without the need for prior models.

\noindent\textbf{3D Object Detection} The final map of Rooms from Motion is equivalent to the output of a standard 3D object detector. However, most 3D object detection frameworks operate on acquired or aggregated point clouds which are sparsely processed by point/voxel-based architectures \cite{rukhovich2022fcaf3d, kolodiazhnyi2024unidet3d} analogous to 2D detection architectures. These methods require access to both pose and depth. Multi-view methods \cite{rukhovich2022imvoxelnet, tu2023imgeonet} relax the depth requirement but generally lose sparse computation by backprojecting rays into a dense 3D volume. As such, they are significantly limited by the memory required to produce the dense volume and scale poorly to small objects or large scenes. Rooms from Motion, however, computes and stores sparse object-centric features independent of image resolution which allows for limited memory footprint.

\section{Rooms from Motion}

\begin{figure}
\centering
\captionsetup{type=figure}
\includegraphics[width=0.98\textwidth]{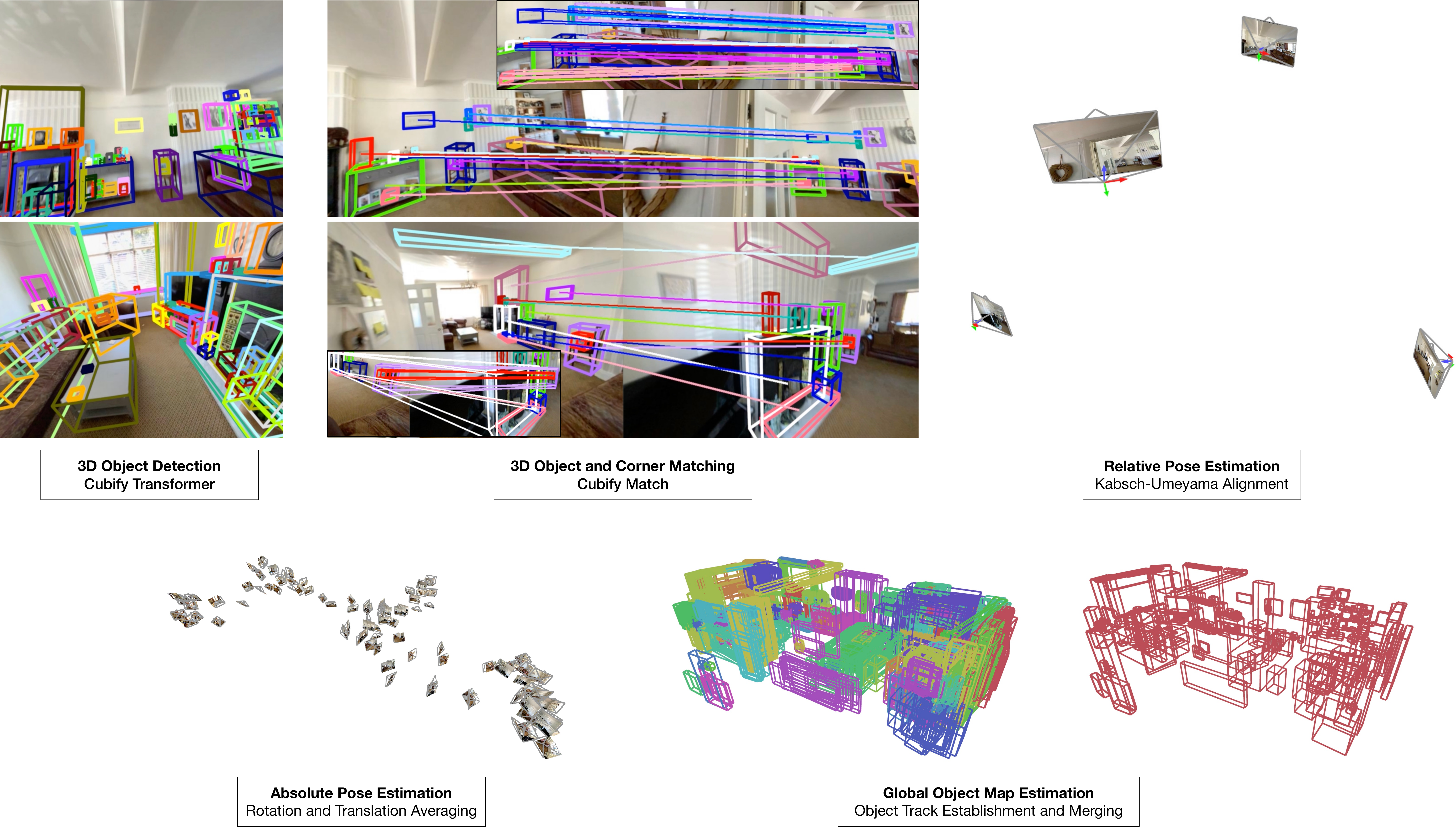}
\caption{An overview of Rooms from Motion. \textbf{Two-view geometry} is considered first. Semantic objects as 3D boxes are independently detected in individual frames using CuTR (Section \ref{sec:detecting_objects}). Objects are subsequently matched at the 3D box level (Sec. \ref{sec:matching_objects}) and at the implicit 3D box corner level (Sec. \ref{sec:match_corners}). Relative pose can be estimated using the matched objects and corners between the images as point sets (Sec. \ref{sec:relpose_estimation}). \textbf{Averaging} operates on the estimated relative poses to produce global camera poses (Sec. \ref{sec:averaging}). Given the set of all matched objects, we establish \textbf{object tracks} and use each observation to aid in estimating the global 3D box and semantics for each track. Finally, duplicate tracks are subsequently merged or suppressed. \textit{Bottom right:} We show lifted 3D box observations belonging to each object track by color as well as the final \textit{representative} 3D boxes (red) for each track.}
\label{fig:two_view_overview}
\end{figure}

Rooms from Motion (RfM) operates on an unordered collection of RGB or RGB-D images with or without \textit{a priori} poses. We describe the components within our framework in an analogous manner to a typical global SfM pipeline \cite{pan2024global}: detection, matching, relative pose estimation, global pose estimation, track formation, and global optimization. We provide a high-level overview of our method in Figure \ref{fig:two_view_overview}. We refer readers to \cite{lazarow2024cubify} for specifics on the Cubify Transformer architecture and \cite{pan2024global, pan2024gravity} for details related to 4-DoF rotation and translation averaging.

\subsection{Detecting 3D objects}
\label{sec:detecting_objects}

Analogous to keypoint detectors \cite{lowe2004distinctive, rublee2011orb, detone2018superpoint}, we rely on a strong image-based 3D object detector capable of predicting oriented 3D boxes. Unlike traditional keypoint-based methods  we require 3D boxes be metrically scaled. Therefore, we adopt Cubify Transformer (CuTR) \cite{lazarow2024cubify} as our base 3D object detector. Given an image $\mathcal{I}$ and optionally a depth map, CuTR produces 3D object detections in the form $\left(\mathcal{B}, \mathcal{F}, \mathcal{C}, \mathcal{S}\right)$ which correspond respectively to 3D oriented boxes, per-object feature embeddings, classification labels, and classification scores. Since CuTR's output is \textit{resolution independent}, we run CuTR on high resolution images (e.g., $876 \times 584$ for ScanNet++) which assists in detecting small and far objects.

\subsection{Matching 3D objects} \label{sec:matching_objects}

Analogous to \textit{2D keypoint} matchers \cite{sun2021loftr, sarlin2020superglue, lindenberger2023lightglue}, we design a \textit{3D object} matcher. We extend CuTR and apply the general design from LightGlue \cite{lindenberger2023lightglue} to objects. Specifically, we consider two images $\mathcal{I}_1$ and $\mathcal{I}_2$. CuTR produce a set of detections in the form of oriented 3D boxes $\mathcal{B}_1$ and $\mathcal{B}_2$ for each image alongside features $\mathcal{F}_1$ and $\mathcal{F}_2$ corresponding to the final activations of the query embeddings. We threshold these sets by $\tau$ (0.25 for CA-1M, 0.2 for ScanNet++) using the confidences in $\mathcal{S}_1$ and $\mathcal{S}_2$.

Given the thresholded objects $\left(\mathcal{B}_1^\tau, \mathcal{F}_1^\tau\right)$ and $\left(\mathcal{B}_2^\tau, \mathcal{F}_2^\tau\right)$, we treat the features $\mathcal{F}_1^\tau$ and $\mathcal{F}_2^\tau$ as the inputs to a sequence of self and bidirectional-attention layers from LightGlue \cite{lindenberger2023lightglue}. We use a positional encoding derived from the respective 3D boxes $\mathcal{B}_1^\tau$ and $\mathcal{B}_2^\tau$. Pairwise object matching scores and a partial object assignment \textbf{P}$_{ij}$ are the output of the network. We define the set of \textbf{matched objects} by those matches in the object assignment whose scores are above a threshold (0.5 for CA-1M and 0.4 for ScanNet++). We generally use a matching network of 3 blocks. We do not use pruning or early exiting since the number of objects in an image is relatively small. We supervise the matching using the ground-truth object IDs in each dataset. Additional details are provided in the supplementary.

\subsection{Matching 3D boxes}
\label{sec:match_corners}

While it is straightforward to define a coarse object matcher as in Section \ref{sec:matching_objects}, we do not immediately have a set of point correspondences which may be used to derive a relative transformation between $\mathcal{I}_1$ and $\mathcal{I}_2$. While one may implicitly derive a point (e.g., the center or corner) from each 3D box to use as a correspondence, objects (unlike points) are often partially observed and these implicitly matched points would be unlikely to correspond to the same point in 3D. Therefore, we design an \textit{additional matching network} of the same design as in Section \ref{sec:matching_objects}, but which operates top-down on points of interest derived from the corners of each 3D box. Specifically, we use the network to match two point sets derived by concatenating the eight box corners $C_i = \{ (x_i^1, y_i^1, z_i^1), \ldots, (x_i^8, y_i^8, z_i^8) \}$ across all objects in each image into a global ``box cloud'' of $8 \times |\mathcal{B}_i|$ queries alongside a positional encoding derived from each corner's 3D position and an interpolated 2D image feature at its 2D projection. We ensure the point-level matches respect the object-level matches by only retaining point-level matches which correspond to the same match at the object level. Additional details are provided in the supplementary.

\subsection{Relative pose estimation and geometric verification}
\label{sec:relpose_estimation}
Given the set of matched objects and corresponding matched 3D corners, we estimate relative pose $R'_{12}$ using Kabsch alignment \cite{kabsch1976solution} with the matched 3D corners between $\mathcal{I}_1$ and $\mathcal{I}_2$ serving as point sets. Since the 3D boxes/corners are expected to already be metrically accurate, the translation components of $R'_{12}$ are also expected to be metric. We treat this as a 4-DoF alignment (yaw and translation) and use circular regression \cite{pan2024gravity} since CuTR already assumes the existence of gravity measurements that are readily available on commodity devices.

Some object and corner matches may be incorrect, therefore, we estimate relative poses across the matched objects using a minimal set of 2 matched objects. We \textit{exhaustively} test these samples since the number of matched objects between two frames is generally small. For each minimal set, we do not perform any sampling on the matched corners. For each matched object pair $(B^1_m, B^2_n)$, we compute a box-based matching error as $1 - IoU3D(B^{12}_m, B^2_n)$ where $B^{12}_m$ corresponds to the reprojection of box $B^1_m$ from image $I_1$ to $I_2$ under the estimated $R'_{12}$. We consider a \textit{matched object} to be an inlier match if the underlying box reprojection error is less than $0.75$ (i.e., $IoU3D \geq 0.25$). We consider a \textit{sample} to be an inlier if each box in the sample is an inlier under $R'_{12}$ and the ratio of inliers when considering the full set of matched objects between $I_1$ and $I_2$ under $R'_{12}$ is at least $0.5$. We take the inlier sample (if any) with lowest mean matching error to produce a verified relative pose $R_{12}$ Additionally, we can compute a corner matching error by taking the $L^2$ distance of the reprojection of the matched corners of each box. We define a \textit{matched corner} to be an inlier match if its reprojection under $R'_{12}$ is less than 10 cm from the matched corner in $I_2$. Corner inliers are used solely within global optimization (Section \ref{sec:ba}).

\subsection{Absolute pose, object track establishment, and merging}
\label{sec:averaging}
\paragraph{Averaging and filtering} We run CuTR on every image and Cubify Match between each pair of images $I_i$ and $I_j$ to estimate relative poses $R_{ij}$ and produce a view graph $G$ consisting of images $I_i$ and geometrically verified $R_{ij}$ relative poses as edges. We use \verb|glomap| \cite{pan2024global} as a guide to perform rotation and translation averaging. Therefore, we run three rounds of rotation averaging and eliminate outlier relative pose estimates which have disagreement with the estimated relative rotations and the relative rotation after averaging by more than 3 degrees. For translation averaging, we fix the scale parameter since we desire metric scale. We run three rounds of translation averaging and remove outlier relative pose estimates whose relative translations disagree by more than 10 cm with the relative translations derived after averaging. After averaging and filtering, we have a global pose ${RT}_i$ for each image $I_i$ within the largest connected component of $G$.

\paragraph{Object track establishment} While a traditional SfM pipeline would form \textit{point} tracks, we initially form \textit{object} tracks. Each object track $O_j$ is characterized by a set of $(I_i, B_i)$ pairs which correspond to an observation of that object in image $i$ as 3D box $B_i$. Each object track is formed by taking the union-find across all inlier object matches remaining after the averaging process. Now, given an object track $O_j$ with a series of observations $(I_i, B_i)$, we can lift each $B_i$ into the global frame using ${RT}_i$. We then must define a \textit{representative} 3D box $\mathcal{B}_j$ for the track. While many possible methods exist, we choose a simple initialization based on the box $B_i$ which has the highest geometric mean of average mutual 3D IoU with all other 3D boxes in the track and underlying detection score. For experiments where semantic classes are available (e.g., ScanNet++ \cite{yeshwanth2023scannet++}), we produce a class distribution weighted by the individual classification scores of each observation and use the class with the maximum probability as the representative semantic label for the track. Finally, we include details on \textit{track merging} in the supplementary.

\subsection{(Partial) Bundle Adjustment: Object track optimization}
\label{sec:ba}

\begin{figure}
\centering
\captionsetup{type=figure}
\includegraphics[width=0.98\textwidth]{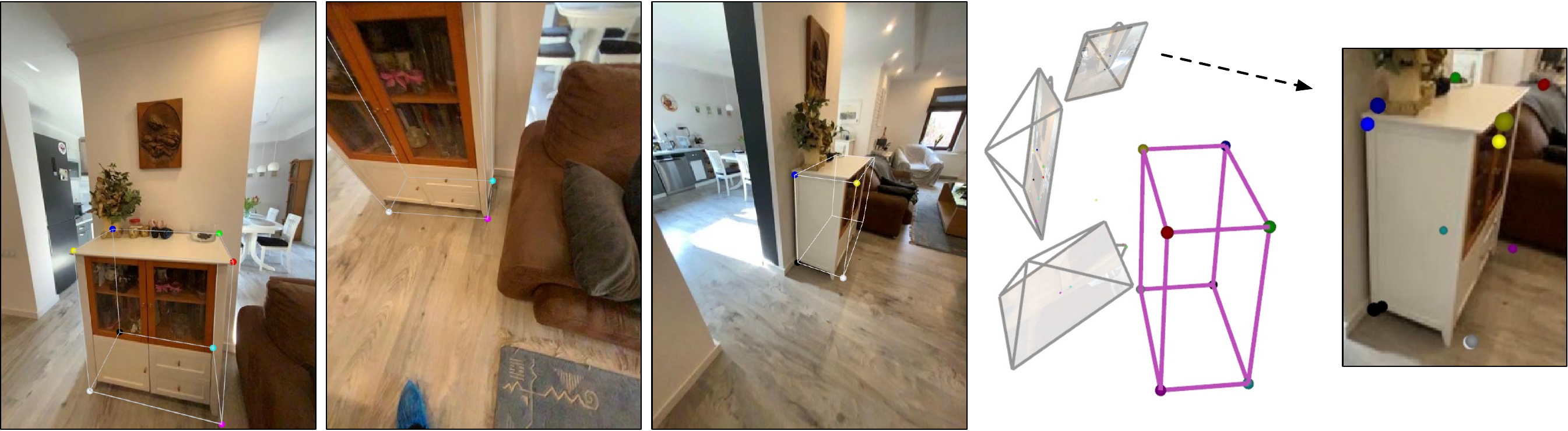}
\caption{\textbf{Bundle adjustment} can be extended to a global 3D object track. Each observation of a track is characterized by an image and particular 3D box detection from CuTR. Since these individual 3D boxes (shown as thin, white boxes) are usually aligned well with the RGB image, they can be used to enforce a reprojection cost function by comparing the projected corners of these detections to the projected corners of the object track's \textit{representative} 3D box. The point tracks established in Section \ref{sec:match_corners} (shown in 8 colors) allow us to map the corners of these individual 3D boxes to the corners of the track's global 3D box. In the rightmost pane, we show an initial mismatch of the projected corners from the individual detection (brighter color) and the object track (darker color) that may be minimized by optimizing the track's 3D box parameters.}
\label{fig:ba}
\end{figure}

Global optimization can further aid in the accuracy of global 3D boxes given the partial or occluded observation that naturally happens within images. We rely on global optimization to improve map quality in the scenario where we have access to pose but lack other metric information (i.e., we use monocular RGB images). Global optimization allows us to optimize the reprojection error of the representative 3D box $\mathcal{B}_j$ to each observation $(I_i, B_i)$. Since optimizing the 3D box as a volume (rendering) is expensive, we choose to use the 3D box corners to establish \textit{point tracks} analogous to the explicit point tracks in traditional SfM. Using a similar procedure to establishing object tracks, we consider the union-find across the 3D corner inlier matches from the representative box $\mathcal{B}_j$ to 3D corners for all other box observations in the object track. This produces point tracks of the form $P_j(n) = \{ (I_i, B_i, m)$ where $1 \leq n, m \leq 8$ corresponding to which corner $m$ of the original image-based box $B_i$ corresponds to corner $n$ of the representative box $\mathcal{B}_j$. We differentiably optimize the underlying 3D box through a cost function that considers the projection of each corner's point track to each observation using PyCeres \cite{Agarwal_Ceres_Solver_2022}. We visualize an example of a single object track problem across three views in Figure \ref{fig:ba}. Additional details are provided in the supplementary.

\section{Experiments}

We evaluate Rooms from Motion on its mapping (i.e., 3D object detection) and metric localization capabilities. We conduct experiments on two large-scale datasets: CA-1M \cite{lazarow2024cubify} and ScanNet++ \cite{yeshwanth2023scannet++}. CA-1M, while class-agnostic, contains exhaustive labeling of objects as oriented 3D box across thousands of captures. This allows us to explore Rooms from Motion's ability to localize and map \textit{any object} regardless of semantics. Additionally, CA-1M has multiple captures of the same room, allowing us to study relocalization. We use ScanNet++ (1K) to study \textit{semantic classification} using its 84-class instance taxonomy. Unlike CA-1M, 3D boxes are not explicitly annotated and axis-aligned boxes are derived from provided instance segmentation. Additionally, the DSLR captures of ScanNet++ allow us the opportunity to study the behavior of collections that are not captured from a continuous trajectory. Both datasets have  access to highly accurate ground-truth poses and ground-truth depth. Therefore, we perform experiments in a few settings:

\begin{enumerate}
    \item \textit{Ground-truth depth and poses}: This explores a pure mapping variant of Rooms from Motion where it does not need to estimate camera poses and where Cubify Transformer has access to ground-truth depth from RGB-D images. This is directly comparable to the most common setting of 3D object detection where a ground-truth point cloud is the input to a sparse 3D object detector \cite{rukhovich2022fcaf3d, kolodiazhnyi2024unidet3d}.
    \item \textit{Ground-truth depth alone:} This explores a localization and mapping variant where the metric pose must be estimated, however, Cubify Transformer has access to strong metric information through ground-truth depth maps. This explores the setting where depth may be available, however, running a distinct localization pipeline may be undesirable.
    \item \textit{Monocular RGB and ground-truth pose:} This also considers the pure mapping variant of Rooms from Motion where Cubify Transformer only has access to monocular RGB images (i.e., without scale information) and must recover scale information. This is directly comparable to \textit{multi-view} 3D object detection methods \cite{rukhovich2022imvoxelnet, tu2023imgeonet}.
    \item \textit{Monocular RGB alone:} This is the most extreme setting where only RGB images are available. Rooms from Motion must estimate metrically accurate camera poses and 3D box predictions purely from RGB images.
\end{enumerate}

\paragraph{Inference settings} While point and volume-based methods are \textit{feedforward} methods which run once on an aggregated point cloud, RfM runs on each image from the corresponding device captures. We uniformly sample all captures into 100 frames. Rooms from Motion is oblivious to temporal aspects (e.g., locality) of this sampling. For CA-1M, we evaluate on the unique set of 3D objects observed across the ground-truth in all sampled frames. For ScanNet++, we always evaluate on the full set of 3D boxes since each capture corresponds to a specifically labeled mesh. This advantages the point-based methods over RfM since there are no guarantees the DSLR capture observes all such objects. We do not evaluate on walls, ceilings, or floor objects.

\paragraph{Un-posed alignment} When evaluating camera poses or 3D object detection results, we align all un-posed results by a \textit{pure rotation and translation} (i.e., $SE(3)$) using the traditional alignment of the positions (or rotations) of each set of camera poses. No scaling is used.

\subsection{Mapping: 3D Object Detection}\label{sec:3dod}

\begin{table*}[ht!]
    \renewcommand{\arraystretch}{1.5}
    \begin{center}
    \caption{We evaluate the mapping ability of Rooms from Motion against state-of-the-art, point-based 3D object detection methods. All methods use RGB images and ground-truth depth and are limited to 1000 detections. Only Rooms from Motion can produce 3D maps without \textit{a priori} poses.}    
    \label{tab:big_table_depth}
    \resizebox{.8\textwidth}{!}{
    \begin{tabular}{|l|cccc|cccc|}
        \hline 
        & \multicolumn{4}{c|}{\textbf{CA-1M}} & \multicolumn{4}{c|}{\textbf{ScanNet++}}  \tabularnewline[0.3em]
        \hline
        Method & AP15 & AR15 & AP25 & AR25 & AP15 & AR15 & AP25 & AR25 \\
        \hline
        \textit{Global, volume-based methods} & \\
        \hline
        FCAF (Posed) \cite{rukhovich2022imvoxelnet} & 28.7 & 32.5 & 24.3 & 28.9 & 28.6 & \textbf{55.7} & 26.8 & \textbf{51.9} \\
        UniDet3D (Posed) \cite{kolodiazhnyi2024unidet3d} & - & - & - & - & 33.3 & 48.1 & 31.6 & 46.0  \\
        \hline
        \textit{Sequential, image-based methods} & \\
        \hline
        Rooms from Motion (Un-posed) & 39.6 & 47.2 & 30.8 & 38.7 & 33.4 & 42.9 & 29.6 & 38.0 \\
        Rooms from Motion (Posed) & \textbf{47.4} & \textbf{53.8} & \textbf{38.9} & \textbf{45.6} & \textbf{38.8} & 48.6 & \textbf{35.6} & 45.0 \\
        \hline
    \end{tabular}}
    \end{center}
\end{table*}
\vspace{-2mm}
\begin{table*}[ht!]
    \renewcommand{\arraystretch}{1.5}
    \begin{center}
    \caption{We evaluate the mapping ability of Rooms from Motion against multi-view 3D object detection methods. All methods use only RGB image.}    
    \label{tab:big_table_multiview}
    \resizebox{.8\textwidth}{!}{
    \begin{tabular}{|l|cccc|cccc|}
        \hline
        & \multicolumn{4}{c|}{\textbf{CA-1M}} & \multicolumn{4}{c|}{\textbf{ScanNet++}}  \tabularnewline[0.3em]
        \hline
        Method & AP15 & AR15 & AP25 & AR25 & AP15 & AR15 & AP25 & AR25 \\
        \hline
        \textit{Global, volume-based methods} & \\
        \hline
        ImVoxelNet (Posed) \cite{rukhovich2022imvoxelnet} & 13.2 & 18.8 & 8.3 & 12.9 & 17.7 & 30.1 & 15.1 & 26.8 \\
        ImGeoNet (Posed) \cite{tu2023imgeonet} & 18.6 & 23.1 & 13.7 & 17.8 & 18.8 & 31.3 & 16.9 & 27.8 \\
        \hline
        \textit{Sequential, image-based methods} & \\
        \hline
        Rooms from Motion (Un-posed) & 17.0 & 25.7 & 10.7 & 17.6 & 18.6 & 26.1 & 13.6 & 19.5  \\
        Rooms from Motion (Posed) & 24.6 & 34.3 & 17.7 & 26.1 & 27.5 & 36.1 & 23.1 & 30.3 \\
        Rooms from Motion (Posed + BA) & \textbf{31.3} & \textbf{40.0} & \textbf{23.2} & \textbf{31.3} & \textbf{30.5} & \textbf{39.3} & \textbf{25.6} & \textbf{33.4} \\ 
        \hline
    \end{tabular}}
    \end{center}
\end{table*}

We present the main 3D object detection results in Tables \ref{tab:big_table_depth} and \ref{tab:big_table_multiview}. We use the standard average precision (AP) and average recall (AR) metrics at IoU 3D levels of 15\% and 25\%. For CA-1M, the metrics are class-agnostic. For ScanNet++, the metrics are averaged across the standard 84 instance classes. Overall, our results support that \textit{Rooms from Motion is a strong 3D object detector in a wide range of configurations}:

\textbf{Ground-truth depth and poses:} In Table \ref{tab:big_table_depth}, we compare to point-based baselines FCAF \cite{rukhovich2022fcaf3d} on both CA-1M and ScanNet++ where ground-truth depth and poses are used to aggregate the requisite input point cloud. On ScanNet++, where meshes are available, we additionally compare to UniDet3D \cite{kolodiazhnyi2024unidet3d} which relies on superpoints. Rooms from Motion significantly outperforms FCAF in this setting despite it operating at a 1 cm voxel resolution. Even the unposed variant of RfM is capable of outperforming FCAF. Since CA-1M consists of many small objects and ground-truth depth may not always be complete due to LiDAR scanner artifacts, we believe that the image-centric nature of RfM allows for significantly improved detection abilities. On ScanNet++, Rooms from Motions achieves significantly improved precision metrics over UniDet3D and FCAF, not only validating its detection capabilities, but also validating its semantic classification quality. While the point-based methods show better recall, especially FCAF which computes a 3D box for every position in the sparse voxel grid, this comes at a cost of decreased precision. Furthermore, we again see that even an unposed variant of RfM can outperform FCAF in average precision.

\textbf{Monocular RGB and poses:} In Table \ref{tab:big_table_multiview}, we compare to multi-view baselines ImVoxelNet \cite{rukhovich2022imvoxelnet} and ImGeoNet \cite{tu2023imgeonet}. These methods only rely on RGB images and ground-truth pose. However, in the absence of depth, they build dense feature volumes and run dense 3D convolutions in order to process the scene in a single pass. Therefore, this also inherently limits them with respect to GPU memory and even on 32 GB V100, they can only use 45 linearly sampled views and a voxel size of 8 cm. This may, in part, explain the significant gaps in all metrics between them and Rooms from Motion. While the unposed variant of RfM is competitive, the posed variant has greatly improved precision and recall across both CA-1M and ScanNet++. Furthermore, optimizing the object tracks through partial bundle adjustment (BA) from Section \ref{sec:ba} significantly boosts map quality and leads to an overall 60-70\% improvement across all metrics compared to ImGeoNet. Not only is the advantage of using \textit{pixels} rather than \textit{voxels} evident from our results, but also compelling is the ability to optimize object tracks with respect to per-frame observations rather than relying on the aggregation of inputs (i.e., points) that has been historically done in the 3D object detection literature.

\subsection{Localization: Camera Pose Estimation} 

While the previous section considered the mapping performance of Rooms from Motion, understanding the camera pose estimation performance allows us to understand the suitability of box primitives for localization. At the same time, good localization is a prerequisite to good mapping.  We calibrate the performance of Rooms from Motion in both the RGB and RGB-D settings on CA-1M and ScanNet++ in Table \ref{tab:big_table_loc}. While CA-1M captures correspond to a smooth trajectory recorded on a handheld device, the DSLR captures from ScanNet++ may not be smooth and can be challenging for traditional SLAM systems. We emphasize large-scale evaluation (50 sequences for ScanNet++ and 107 for CA-1M) rather than individual sequences. For RGB-D, we compare against DROID-SLAM \cite{teed2021droid} and for RGB, we compare against CUT3R \cite{cut3r} (not to be confused with CuTR). CUT3R is particularly suitable since it maintains \textit{metric} predictions using Mast3r \cite{leroy2024grounding}, which is not always the case \cite{duisterhof2024mast3r, murai2024mast3r}. 

For Rooms from Motion and CUT3R, we subsample uniformly to 100 frames. For DROID-SLAM, we consider denser subsampling at 250 and 500 frames for fairer comparison and this often exceeds the available frames in ScanNet++ DSLR captures. We find ORB-SLAM \cite{mur2017orb} unable to provide reliable tracking and therefore do not compare to it. Since Rooms from Motion only registers the largest connected component of its view graph, we make note of how many frames were, on average, registered per sequence.

\begin{table*}[ht!]
    \renewcommand{\arraystretch}{1.5}
    \begin{center}
    \caption{We evaluate the localization ability of Rooms from Motion. We report both the averaged median error and \textit{(averaged RMSE error)} for rotation and translation as well as the percentage of images that are successfully registered in the global maps. Ground-truth alignments are done with an $SE(3)$ transformation (i.e., no scaling). We note that Rooms from Motion has access to local camera pitch and roll through gravity measurements.}  
    \label{tab:big_table_loc}
    \resizebox{.98\textwidth}{!}{
    \begin{tabular}{|l|ccc|ccc|}
        \hline
        & \multicolumn{3}{c|}{\textbf{CA-1M}} & \multicolumn{3}{c|}{\textbf{ScanNet++}}  \tabularnewline[0.3em]
        \hline
        Method & ARE (deg) & ATE (cm) & Registered & ARE (deg) & ATE (cm) & Registered  \\
        \hline
        DROID-SLAM \cite{teed2021droid} (250 frames, w/ GT depth) & 10.1 (\textit{14.5}) & 14.9 (22.5) & 100\% & 62.8 (\textit{81.0}) & 110.1 (158.8) & 100\% \\
        DROID-SLAM \cite{teed2021droid} (500 frames, w/ GT depth) & 2.6 (\textit{6.1}) & 4.7 (7.8) & 100\% & 58.2 (\textit{79.3}) & 104.7 (\textit{151.2}) & 100\% \\        
        Rooms from Motion (w/ GT depth) & 1.8 (\textit{4.3)} & 4.0 (\textit{8.0}) & 96\% & 1.1 (\textit{2.1}) & 3.5 (\textit{6.7}) & 99.5\% \\
        \hline
        CUT3R (RGB) \cite{cut3r} & 10.7 (\textit{12.6}) & 35.9 (\textit{45.6}) & 100\% & 7.4 (\textit{11.4}) & 27.2 (\textit{36.6}) & 100\% \\
        Rooms from Motion (RGB) & 2.5 (\textit{5.8}) & 12.7 (\textit{17.9}) & 93\% & 2.4 (\textit{5.1}) & 15.3 (\textit{21.7})  & 97\% \\       
        \hline
    \end{tabular}}
    \end{center}
\end{table*}

\vspace{-3mm}
\paragraph{CA-1M} Both Rooms from Motion and DROID-SLAM perform well with handheld RGB-D captures on CA-1M and are able to achieve similar median trajectory error of 4.0 vs. 4.7 cm. With respect to RGB only, CUT3R generally results in significantly worse rotation and translation errors than Rooms from Motion. Surprisingly, Rooms from Motion on RGB (i.e., no depth) is competitive with RGB-D DROID-SLAM running on 250 frames.

\paragraph{ScanNet++} The DSLR captures of ScanNet++ are sparse (usually a few hundred frames) and may not correspond to a smooth trajectory. For SLAM-based methods like DROID-SLAM \cite{teed2021droid}, we can observe degraded performance that likely can be attributed to this despite having access to GT depth. On the other hand, the RGB-D variant of Rooms from Motion is able to register nearly every frame while maintaining a median rotation error of nearly a single degree and 3.5 cm of median translation error. With respect to RGB only, Rooms from Motion continues to significantly outperform CUT3R.

\textit{Rooms from Motion is competent at metric camera localization}. Both DROID-SLAM and CUT3R rely on point-based primitives for localization. Despite relying on 3D box primitives alone, Rooms from Motion registers nearly all frames and provides superior localization ability within the indoor setting alongside sparse captures. We believe this supports the suitability of 3D box primitives as an alternative to point-based approaches for localization.

We perform a basic study of \textit{relocalization} in the supplementary. Additionally, we provide an ablation on how object richness and field of view impact localization ability in the supplementary.

\subsection{Qualitative Map Results}

\begin{figure}[h!]
\centering
\captionsetup{type=figure}
\includegraphics[width=0.98\textwidth]{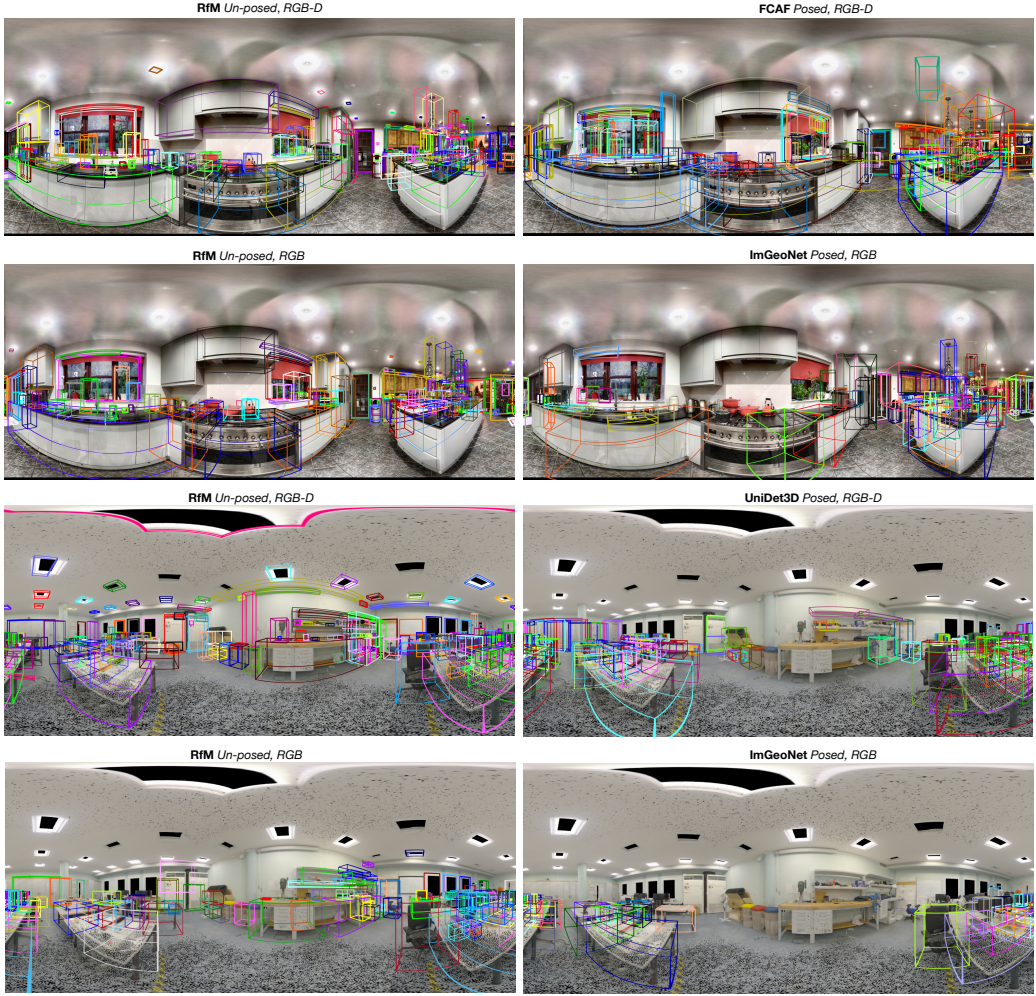}
\caption{Qualitative comparisons from CA-1M (top half) and ScanNet++ (bottom half) of resulting 3D object maps across different methods as projected to the viewpoint of a FARO scanner. We note that RfM results here are \textit{unposed} (i.e., using aligned, estimated pose), while the 3D object detection-based methods rely on \textit{a priori} pose.}
\label{fig:qual}
\end{figure}

We present qualitative 3D object maps across all methods in Figure \ref{fig:qual}. The 3D object maps are projected (not rendered) and overlaid onto a panorama with respect to a FARO stationary scanner position as provided by both CA-1M and ScanNet++. For CA-1M, this is an RGB panorama captured by the scanner software itself, whereas for ScanNet++, this is rendered from the provided mesh. For understanding spatial accuracy, which this projection does not necessarily capture, we refer the reader to the quantitative results of Section \ref{sec:3dod}. For brevity, we only present \textit{unposed} results of Rooms from Motion whereas all other methods are posed.

Rooms from Motion appears to provide the most visually complete maps. Small, thin objects like ceiling lamps and objects on the walls are more faithfully captured within RfM maps. Despite being unposed and viewed from a ``novel'' position (the FARO scanner), we also observe satisfactory RGB alignment as the 3D boxes generally project to the expected object boundaries in the panorama. Additionally, we see that multi-view methods like ImGeoNet particularly struggle when it comes to detecting smaller objects and providing accurate 3D box orientation.

\vspace{-2mm}

\section{Limitations and Discussion}

We introduced a framework which is capable of producing global 3D box detections (i.e., maps) of indoor spaces while simultaneously able to estimate camera poses from unordered RGB(-D) collections. This allows new settings e.g., 3D object detection from un-posed RGB images that may enable more general 3D object mapping of the world.

Our method is currently limited to indoor scenes by a lack of training data of more general indoor and outdoor scenes. At the same time, we can only register images which \textit{have objects} in them. Similarly, more diverse cameras would facilitate the generalization of CuTR and RfM.

\section{Acknowledgements}

We thank Shih-Yu Sun and Sam Parsa for motivating discussions on the intersection between 3D object detection and SLAM. We thank David Griffiths for feedback on the manuscript and helpful discussions. Selected visualizations produced with Rerun \cite{RerunSDK}.

\medskip

{
\small
% ---- Bibliography ----
\bibliographystyle{splncs04}
\bibliography{references}
}

\clearpage
\appendix

\section{\LARGE Appendix}

\subsection{Additional Experiments}
\subsubsection{Ablations}
\label{sec:ablations}

We explicitly ablate two aspects of Rooms from Motion. The first tests how completeness of the objects in the dataset affects the localization performance within RfM. In other words, if we have a larger range of objects with 3D boxes in our dataset, then we should be able to detect more objects within Cubify Transformer and subsequently match more objects within Cubify Match. Secondly, we test how field of view impacts the ability to detect more objects and match more objects between frames.

\begin{table*}[ht!]
\begin{minipage}{.47\linewidth}
\centering
\caption{\textbf{Taxonomy sensitivity} of Rooms from Motion is examined with respect to localization performance using different object subsets of each dataset.}
\label{tab:ablate_taxonomy}
\renewcommand{\arraystretch}{1.5}
\resizebox{.95\textwidth}{!}{
\begin{tabular}{|l|ccc|}
    \hline
    Taxonomy & ARE (deg) & ATE (cm) & Registered \\
    \hline
    ScanNet++ (Standard) & 2.5 & 15.3 & 93\% \\
    ScanNet++ (All) & 2.4 & 15.3 & 97\% \\
    \hline        
    ARKitScenes &  4.4 & 16.2 & 61\% \\
    CA-1M & 2.5 & 12.7 & 93\% \\
    \hline
\end{tabular}}
\end{minipage}\hfill
\begin{minipage}{.47\linewidth}
\centering
\caption{\textbf{Field of view sensitivity} of Rooms from Motion is examined using both the standard FoV images of each dataset versus the additional high FoV captures available in each dataset.}
\label{tab:ablate_fov}
\renewcommand{\arraystretch}{1.5}
\resizebox{.95\textwidth}{!}{
\begin{tabular}{|l|ccc|}
    \hline
    Field of view & ARE (deg) & ATE (cm) & Registered \\
    \hline 
    ScanNet++ (Standard) &  6.1 & 31.0 & 87\% \\
    ScanNet++ (DSLR) & 2.4 & 15.3 & 97\% \\
    \hline        
    CA-1M (Standard) &  7.9 & 26.1 & 68\% \\
    CA-1M (Ultrawide) & 2.5 & 12.7 & 93\% \\        
    \hline
\end{tabular}}
\end{minipage}
\end{table*}

\paragraph{Do more objects help?}
The ability for Rooms from Motion to accurately estimate relative poses between image pairs is inherently limited by the matching capabilities of Cubify Match, which we expect to be highly dependent on \textit{number of objects} that can be detected by Cubify Transformer. Therefore, in Table \ref{tab:ablate_taxonomy}, we ablate the performance of Rooms from Motion with respect to \textit{completeness} of the object labels. For CA-1M, we consider a subset of the object labels corresponding to: 18 types of objects corresponding to ARKitScenes versus the exhaustive CA-1M set of objects. For ScanNet++, we consider the standard 84-class instance taxonomy as well as an expanded taxonomy where we add all remaining instances as ``unlabeled''.

We observe that a coarse taxonomy like ARKitScenes (18 types of objects) appears to be insufficient. Not only are the rotation and translation errors significantly higher than when using CA-1M, but the registration rate significantly drops. Having more objects detected per-frame likely increases the chances of \textit{matching} more objects between frames --- critical for accuracy and connectivity of the view graph.

On ScanNet++, for which the standard taxonomy is 84 types of objects, we observe reasonably good performance. However, we observe improvements in the registration rate (from 93\% to 97\%) when using \textit{all available objects}. Notably, since these remaining objects are added as ``unlabeled'' rather than their specific labels, we see both with ScanNet++ and CA-1M that the localization within Rooms from Motion is not sensitive to having the actual class label of an object.

\paragraph{Does field of view matter?}
All experiments so far use images that correspond to the higher field of view (FoV) images present in CA-1M and ScanNet++. We validate the hypothesis that a larger field of view corresponds to better localization in Table \ref{tab:ablate_fov}. All localization metrics are significantly improved across each dataset alongside large increases in the registration success rate.

\subsubsection{Relocalization} 

We additionally examine the basic ability of Rooms from Motion to \textit{relocalize} given a map. Using CA-1M, which includes 3 captures of each room, we run Rooms from Motion on one of the captures (107 total), and store each registered camera pose alongside its corresponding CuTR detections. Given an additional capture of the same room (222 total), we attempt to relocalize each new image. We run Cubify Match on the new image against the previously registered images. If successful, we use the registered camera pose and estimated relative pose of the image with the most object inliers as the camera pose for the new image. Only a single registered image is selected --- no averaging across relative pose estimates is applied.

\begin{table*}[ht!]
    \renewcommand{\arraystretch}{1.5}
    \begin{center}
    \caption{\textbf{Relocalization performance} of Rooms from Motion on CA-1M. We relocalize additional captures of the same room against an initial Rooms from Motion map of the room.}    
    \begin{tabular}{|l|ccc|}
        \hline
        Input & ARE (deg) & ATE (cm) & Relocalized \\      
        \hline
        RGB & 3.1 (\textit{6.8}) & 15.9 (\textit{24.4}) & 98\% \\
        RGB-D & 2.6 (\textit{4.8}) & 6.0 (\textit{11.2}) & 99\% \\
        \hline
    \end{tabular}
    \end{center}
    \label{tab:reloc}
\end{table*}

We observe high recall with respect to relocalization registration and achieve similar accuracy in rotation and translation errors on par with our full localization experiments in Table \ref{tab:big_table_loc}. Further work remains to understand how to  update the mapping components of Rooms from Motion given additional images.

\section{Additional Model Details}
We provide additional details on our models and setup.

\subsection{General}

All models use the ViT-B variant of Cubify Transformer. Cubify Match is composed of two matching networks of the LightGlue design \cite{lindenberger2023lightglue}, each consisting of 3 self/bi-directional attention layers using a 256-dimensional embedding dimension. All models are trained in PyTorch using a batch size of 64 over 4 x 32GB V100s for the standard schedule from Cubify Transformer \cite{lazarow2024cubify}.

\subsection{Cubify Match}

\subsubsection{More details}

While we generally follow the design of LightGlue and apply it to object (boxes) and points, we more explicitly describe the function.

\paragraph{Object Matching Network}

Given a pair of images, we run Cubify Transformer on each image independently. For CA-1M, we threshold detections by 0.25 (since CA-1M is class agnostic) and for ScanNet++, we threshold by 0.20 (since ScanNet++ is class specific). Each detection includes a 256 dimensional embedding corresponding to the (final) query embedding. Each detection's 3D box is in camera space, we use a small MLP to produce a positional encoding to XYZ center and WLH dimensions (we do not encode the angle). The resulting sets (one for each image) of query embeddings and positional encodings are fed into 3 blocks consisting of self attention and bi-directional attention \cite{lindenberger2023lightglue}. No early exiting/pruning is used, which keeps the network quite simple. Finally, a \href{https://github.com/cvg/LightGlue/blob/main/lightglue/lightglue.py#L273C34-L274C1}{match assignment} must be produced using double softmax which corresponds to an $(N + 1) \times (M + 1)$ similarity matrix of scores where $N$ is the number of thresholded detections in the first image, and $M$ is the number in the second image. The additional row and column correspond to the matchability score which modulates the entire row/column. We use the same \href{https://github.com/cvg/LightGlue/blob/main/lightglue/lightglue.py#L295}{filtering code} in order to produce the final matches (if any).

\paragraph{Box/Corner Matching Network}

Giben the thresholded detections, we project the corners of each 3D box to its respective image to produce 8 $(x, y)$ positions for each box. We bilinearly interpolate the output features of the ViT backbone from Cubify Transformer. We add this to the final query embedding of the corresponding object. We embed the 3D $XYZ$ positions using an MLP to act as a positional embedding. The resulting set of \textit{flattened} inputs (i.e., the inputs are the embeddings/positional encodings of the flattened corners across \textit{all} detections). This essentially gives two ``box clouds'' (i.e., $N \times 8$ points or $M \times 8$ points based on the aggregate of the box corners) as inputs into the matching network. These are fed into 3 blocks consisting of self attention and bi-directional attention \cite{lindenberger2023lightglue}. A match assignment is produced similarly as for the object network using the same thresholds. Given the assignment of corners, we reject assignments of corners which do not correspond to the set of matches from the object matching network.

\subsubsection{Track merging} 

Despite the track establishment process, we may find two object tracks which correspond to the same underlying object but are not connected during establishment when no inlier path between the observations can be established. However, given the existence of a global set of estimated poses, we can consider a merging and suppression procedure. We enumerate all pairs of established object tracks $(\mathcal{O}_i, \mathcal{O}_j)$ and ignore pairs whose generalized IoU 3D is low (smaller than -0.6). Then, we compute the \textit{track affinity} between the tracks by considering the average \textit{observation affinity} between all pairs of observations between the tracks. We define the observation affinity as the product of the object matching score and the generalized IoU 3D of the corresponding representative 3D boxes (shifted by 1.0 to ensure non-negative values). We merge a track when the track affinity exceeds $0.25$. We \textit{suppress} a track when the track does not meet the affinity threshold for merging, but the IoU 3D between the tracks is above $0.15$.

\subsubsection{(Partial) Bundle Adjustment}

Bundle adjustment relies on building point tracks based on the 3D box corners of object tracks. Subsequently, we can define a simple cost function that rather than optimizing the 3D point positions independently, optimizes the underlying box parameters (center, dimensions, yaw) differentiably through the corners. This can be seen in the following code:

\begin{lstlisting}[language=Python, caption=Bundle adjustment cost]
class ReprojectionError(object):
    def __init__(self, size, K, idx, xy):
        # Image size.
        self.size = torch.tensor(size).to(device=xy.device, dtype=torch.float64)
        # Intrinsics.
        self.K = K
        # 3D corner to be projected.
        self.idx = idx
        # Expected 2D projection.
        self.xy  = xy

    # Box parameters and camera pose.
    def __call__(self, center, dims, yaw, rq, T):
        R = quaternion_to_rotation_matrix(rq[None])[0]
        box_xyz = box_to_corners(center, dims, yaw)[self.idx]
        box_xy = self.K @ (R @ box_xyz + T)
        box_xy = box_xy[:2] / box_xy[2:]

        return (box_xy - self.xy) / self.size
\end{lstlisting}

\end{document}